\documentclass[a4paper, 10pt, conference]{IEEEtran}
\IEEEoverridecommandlockouts
\usepackage{cite}

\usepackage{multicol}
\usepackage{graphicx}
\usepackage{color}
\usepackage{amsmath}
\usepackage{multirow}
\usepackage{booktabs}
\usepackage{url}
\usepackage{svg}
\usepackage{amsmath}
\usepackage{adjustbox}
\usepackage{subfloat}
\usepackage[english]{babel}
\usepackage{comment}
\usepackage{tikz}
\usetikzlibrary{positioning}

\begin{document}

\author{
 \IEEEauthorblockN{Navid Ashrafi$^{1,2}$, Vera Schmitt$^2$, Robert P. Spang$^2$, Sebastian M\"oller$^{2,3}$, Jan-Niklas Voigt-Antons$^{1}$}
 \IEEEauthorblockA{$^1$Immersive Reality Lab, University of Applied Sciences Hamm-Lippstadt, Lippstadt, Germany\\
$^2$Quality and Usability Lab, TU Berlin, Germany\\
$^3$German Research Center for Artificial Intelligence (DFKI), Berlin, Germany}
}

\title{Protect and Extend - Using GANs for Synthetic Data Generation of Time-Series Medical Records\\
{\footnotesize}
}

\IEEEoverridecommandlockouts
\IEEEpubid{\makebox[\columnwidth]{979-8-3503-1173-0/23/\$31.00
\copyright 2023 European Union \hfill} \hspace{\columnsep}\makebox[\columnwidth]{ }}
\maketitle

\begin{abstract}
Preservation of private user data is of paramount importance for high Quality of Experience (QoE) and acceptability, particularly with services treating sensitive data, such as IT-based health services. Whereas anonymization techniques were shown to be prone to data re-identification, synthetic data generation has gradually replaced anonymization since it is relatively less time and resource-consuming and more robust to data leakage. Generative Adversarial Networks (GANs) have been used for generating synthetic datasets, especially GAN frameworks adhering to the \textit{differential privacy} phenomena.
This research compares state-of-the-art GAN-based models for synthetic data generation to generate time-series synthetic medical records of dementia patients which can be distributed without privacy concerns. Predictive modeling, autocorrelation, and distribution analysis are used to assess the Quality of Generating (QoG) of the generated data. The privacy preservation of the respective models is assessed by applying membership inference attacks to determine potential data leakage risks. Our experiments indicate the superiority of the privacy-preserving GAN (PPGAN) model over other models regarding privacy preservation while maintaining an acceptable level of QoG. The presented results can support better data protection for medical use cases in the future.
\end{abstract}
\newcommand\copyrighttext{%
  \footnotesize \textcopyright 2023 IEEE. Personal use of this material is permitted. Permission from IEEE must be obtained for all other uses, in any current or future media, including reprinting/republishing this material for advertising or promotional purposes, creating new collective works, for resale or redistribution to servers or lists, or reuse of any copyrighted component of this work in other works. Navid Ashrafi, Vera Schmitt, Robert P. Spang, Sebastian Möller, and Jan-Niklas Voigt-Antons. 2023. Protect and Extend - Using GANs for Synthetic Data Generation of Time-Series Medical Records. In 2023 15th International Conference on Quality of Multimedia Experience (QoMEX), Ghent,Belgium,2023, pp. 171–176. https://doi.org/10.1109/QoMEX58391.2023.10178496. https://ieeexplore.ieee.org/document/10178496}
  
\newcommand\copyrightnotice{%
\begin{tikzpicture}[remember picture,overlay,shift={(current page.south)}]
  \node[anchor=south,yshift=10pt] at (0,0) {\fbox{\parbox{\dimexpr\textwidth-\fboxsep-\fboxrule\relax}{\copyrighttext}}};
\end{tikzpicture}%
}
\copyrightnotice

\begin{tikzpicture}[overlay, remember picture]

\path (current page.north) node (anchor) {};

\node [below=of anchor] {%

2023 15th International Conference on Quality of Multimedia Experience (QoMEX)};

\end{tikzpicture}

\section{Introduction}
The current General Data Protection Directive (GDPR) limits access and distribution of personal medical records, making it difficult to develop applications and train deep learning models due to restricted data access. Generating high-quality synthetic data is a necessary requirement for the improvement of patients' Quality of Experience (QoE). Publicly available data for training interactive systems in the medical domain is scarce due to security and privacy reasons. Whereas this data is needed for training algorithms that enable adaptation of the user interface and content to match the preference and abilities of users, all of which would directly contribute to an improved patient's QoE. For data-driven approaches in the medical domain, having sufficient training data while maintaining anonymity and data utility is crucial. This is especially true for deep learning models used in domains like natural language processing and information retrieval, where a substantial amount of training data is a basic requirement \cite{torfi2022differentially}. Although a significant amount of data is available in healthcare, it cannot be utilized for machine learning applications due to the GDPR restrictions. This limits the deployment of deep learning models in healthcare \cite{guan2018generation}.

Anonymizing medical records has been proposed as a solution to this problem and has become a popular method for making real medical records available for public research. However, many of these approaches have proven to be fragile against re-identification attacks. Synthetic data generation by applying deep learning models such as Generative Adversarial Networks (GANs) offers higher data privacy \cite{DGenEvaluation}. GAN models for synthetic text data are rapidly evolving to meet the demand for privacy-preserving data sharing in research. However, classical GANs have the ability to remember training samples, making it possible to re-identify personal-related information. Other approaches such as diffusion models \cite{dif} have been suggested as potential alternatives for GANs in synthetic data generation while privacy issues are not addressed for this approach and the main focus has so far stayed on the data utility. Hence, we will use GANs for our use case. 

Recent GAN-based frameworks have introduced innovative ideas to improve the quality of generated data and enhance privacy at the same time, such as integrating privacy-preserving mechanisms. However, generating synthetic data is challenging due to the data utility and privacy trade-off. The privacy-utility trade-off refers to the idea that in certain situations, increasing the privacy of data generation may come at the cost of reducing the quality. Achieving an effective balance between the two factors can be a challenging task, as finding suitable GAN approaches is not a straightforward process \cite{wang2020part}. 
%In GAN-based synthetic data generation, the most significant hurdles are ensuring a high level of quality in the generated data and producing a distribution that is comparable to the original data, all while addressing privacy concerns.

The main research question of this contribution is: Can high-quality synthetic datasets for time series medical records be generated while safeguarding privacy?
In this paper, various GAN-based models are utilized to explore their performance with respect to data utility and privacy. Five GAN-based data generation frameworks have been trained, namely simpleGAN, medGAN, DoppelGANger (DG), Differentially Private GAN (DPGAN), and Privacy Preserving GAN (PPGAN). The objective of using a dementia patients' dataset to generate synthetic data is to create new data with a high quality of generation (QoG), while also ensuring strong privacy protection. This privacy-preserving high-quality data can also be used for \emph{extending} existing datasets to foster medical research for machine learning applications. 

%Often there is a lack of data to train and fine-tune deep learning models to improve services, applications, and their Quality of Experience. For example, it is necessary to have sufficient training data for algorithms that are taking decisions for adaption strategies of how to deliver the user interface or how to manipulate interaction strategies in medical applications.

The upcoming sections will provide an overview of the related work (Section \ref{rw}), followed by a description of pre-processing of the medical records and GAN-based models used for the synthetization (Section \ref{methodology}). Subsequently, in Section \ref{results} results of the performance of different GAN-based frameworks are compared, followed by a discussion (Section \ref{discussion}) of the findings and concluding remarks (Section \ref{conclusion}).

\section{Related work} \label{rw}
\subsection{Data Anonymization}
Legal regulations, such as the GDPR, require the anonymization of personal data. However, clear guidelines are still lacking facilitating the development of precise and uniform approaches to preserve the privacy of personal information \cite{olatunji2022review}. Overall, anonymization refers to the exclusion or removal of personal identifiers. Furthermore, anonymization approaches require manual manipulation of data identifiers whereas, in the case of extensive datasets, anonymization remains prone to human error and re-identification attacks.

\subsection{Synthetic Data Generation}
Synthetic data generation emerged to mitigate the re-identification of individuals. Hereby, most of the machine learning frameworks, such as Autoregressive models (AR), Recurrent Neural Networks (RNN), and Variational Autoencoder (VAE) used for the synthetic data generation approach are data-type specific \cite{buczak}. However, most models often do not consider metadata and time series sequences, failing to capture complex temporal dependencies. Whereas, in the healthcare domain, most datasets contain time series and sequences. Hereby, Anonymization and ML-based generative models for synthetic medical data generation did not provide ideal data utility and privacy. Thus, GAN-based approaches have been proven to outperform VAE, RNN, and AR for the synthetic data generation task. 

\subsection{GAN-based Approaches}
One of the earlier contributions applying GANs for synthetic medical records generation is medGAN \cite{medGAN}. This framework cannot sample time series and sequential data and considers the whole dataset as one large matrix. However, medGAN motivated the development of other GAN frameworks capable of handling discrete and continuous features. Still, these models do not address privacy-preserving mechanisms but yield a good QoG performance \cite{torfi2022differentially}. One approach addressing privacy preserving \textit{deep learning} mechanisms to generate synthetic data from labeled but also unlabeled data is PATE-GAN \cite{jordon2018pate}. This approach uses \textit{Private Aggregation of Teaching Ensembles} to ensure differential privacy \cite{torfi2022differentially}. A further model developed based on medGAN is ADS-GAN \cite{paper5} which manipulates the classic GAN architecture to improve the QoG by feeding additional input to the GAN's generator and discriminator. Additionally, \textit{ADS-GAN} uses Wasserstein GAN with Gradient Penalty (WGANGP) \cite{wsimproved} to mitigate the mode collapse problem \cite{paper5}. Despite the improved QoG and privacy robustness compared to \textit{WGAN} and medGAN, ADS-GAN is still not designed to capture sequential and time series data distribution. 
A more recent approach for synthetic data generation is DG \cite{DoppleGANger}, which utilizes GANs to enable the synthetic generation of complex, sequential data sets. DG is a generic framework that can take data samples from various datasets and still achieve a high QoG. DG can generate data features conditioned on data attributes, handle extreme events, and prevents mode collapse \cite{DoppleGANger}. Concerning \textit{differential privacy}, DG provides the possibility of including DPGAN in their training process, to add noise to the training samples. However, the added noise results in a drastic reduction of the QoG.  

\subsection{Privacy-preserving GAN-based frameworks}
Algorithms in the domain of \textit{differential privacy} have been widely studied and have been used for data release \cite{zhang2017privbayes}, classification \cite{chaudhuri2011differentially} and deep learning \cite{abadi2016deep}. Various models using the WGAN framework have been proposed. Among those are \textit{differential privacy} GAN (DPGAN) \cite{paper6} and privacy-preserving GAN (PPGAN) \cite{paper2} showing good performance in protecting generated data from re-identification. In the domain of health DPGAN achieves \textit{differential privacy} by adding carefully designed noise to gradients during the learning procedure, whereas PPGAN combines well-designed noise with training gradients to disturb the distribution of the original data.

\subsection{Data Security} \label{mia}
Membership inference is a common way to evaluate the re-identification of individuals and evaluate the privacy-preserving ability of different models \cite{chen2020gan}. The goal of such attacks is to infer if specific data samples were used in the training process of a particular ML model. This approach introduced a data re-identification method where an adversary trains a set of shadow models replicating a target model $M$ \cite{shokri}. Using the outcome of these shadow models, the adversary can then train an attacker model $M_{attack}$ that would be capable of determining whether or not a data sample \textbf{X} has been used to train $M$ \cite{shokri}. \textit{TensorFlow Privacy} \cite{tensorFlow4} has introduced a package containing a collection of different membership attacks (\textit{TensorFlow Privacy}). Unlike the classic membership inference attacks, this approach does not require the shadow model since the target model is being used to serve as a shadow model. The model offers a data-slicing mechanism where separate attacks are executed on various chunks of the dataset. Also, Various types of attacks are included in \textit{TensorFlow Privacy}. After empirically testing different attacker types, we found that \textit{Logistic Regression} and the \textit{Threshold Attack} achieved the best performance for the given dataset.

\section{Methodology} 
\label{methodology}
\subsection{Dataset: time series medical record} 
The dataset used in this work contains time series medical records of dementia patients\cite{pflegetab}. The data contains the patient's interactions with a tablet game (\textit{PflegeTab}), where patients are asked to fulfill specific tasks in the game. The tablet game was used to rate patients' \textit{Quality of Life} (QoL) by collecting in-game touchscreen behavior. Additionally, the dataset contains further metadata such as time spent on a specific task, cognitive impairment severity, and overall interactions with the game. The dataset contains interactions from 81 users over 936 days and an overall playtime of $193.72h$ and $33.377$ completed tasks.
The datasets were pre-processed for each task/experiment based on its requirements. We decided to keep the date feature representing the date on which, a user used the game in our original dataset. However, we did not need to include the date column in the generation process since it is not considered an important feature. Similarly, the \textit{user\_id} representing the ID number of each user was also excluded from the dataset after creating the sequences. 
Several data pre-processing steps were conducted to prepare the data for the GAN-based models. The \textit{user\_mmst} values representing the Mini-Mental-Status-Test (MMST) scores for each user, have been taken to generate sequences. Then for each dataset, we stacked all the sequences on top of each other to have a more organized dataset, and we clipped the sequences that were larger than a threshold (maximum sequence length). We configured the threshold for each data empirically based on their sequence length to keep all the datasets in a reasonably similar size (5K-8K rows), otherwise, the comparison would not yield fair results. 
In some autocorrelation evaluation steps, we needed to represent each patient in one row of data. To accomplish this, we calculated the mean value for each column in each sequence and returned it as a single value for the respective column. As a result, we had a small dataset of 56 rows (one row per user) in which any value represented the mean value of all rows for a specific patient.
For the evaluation, if the prediction performance increases, when the original dataset is extended by the synthetically generated data, the synthetic data, and real data are merged. 

\subsection{GAN frameworks for synthetic data generation} 
In the following, the GAN-based models used for the synthetic generation task are described. 

\textbf{simpleGAN:} the simpleGAN framework has been used with a sequential Keras model dividing the data into \textit{test} and \textit{train} subsets by a 7:3 ratio. We used \textit{10K} epochs to train the GAN engine containing a discriminator, a generator, and the GAN functions. We generated \textit{10K} rows of synthetic data using the trained generator function. As simpleGAN transforms high-dimensional complex distributions into low-dimensional simple distributions, in the following experiments, the model is used as a baseline comparing the QoG with other GAN-based models. We kept the same number of rows (10K) when generating synthetic datasets using all of the upcoming frameworks.

\textbf{medGAN:} This framework is one of the first approaches in the medical domain introducing a GAN that generates high-dimensional, multi-label discrete variables that represent events in electronic health records \cite{medGAN}. Although medGAN provides detailed guidelines on how to utilize the model in terms of pre-processing and training, we implemented our own pre-processing steps for the model since our dataset contains a different structure, and therefore, the suggested pre-processing steps could not be followed. 

\textbf{DoppleGANger:} DG is a model leveraging GANs for generating synthetic data from time series data, which is a prevalent data type usually in domains such as finance, networking, health, and many other fields \cite{DoppleGANger}. DG includes several improvements, such as adding an LSTM including a batch setup which is designed such, that each LSTM cell outputs multiple time points to improve temporal correlations and synthesize long time series that are representative. To reduce mode collapse and improve the training process DG applies the Wasserstein loss with gradient penalty (WGAN-GP) \cite{wsimproved}. The modification implemented in DG to handle time series records and improvements made on the basic GAN architecture, lead to an expressive time series model with the ability to produce high-fidelity synthetic time series data. Therefore, DG is implemented for the dementia patients' data to examine its performance in comparison with the other GAN-based models described in this chapter. 

\textbf{DPGAN:} The DPGAN framework attempts to preserve privacy during the training procedure by adding noise to the gradients of the \textit{Wasserstein} distance in the discriminator concerning the sample training data \cite{paper6}. DG integrated DPGAN into their framework to generate differentially private data samples. However, DPGAN can be easily adopted also for other GAN-based frameworks and does not rely on WGAN only. The post-processing property of \textit{differential privacy} states that any mapping after \textit{differential privacy} has been applied will not invade privacy \cite{paper6}. The mapping in DPGAN is the computation of parameters of the generator, and the \textit{differential private} parameter of the discriminator \cite{paper6}. DPGAN is integrated into the DG framework in this analysis as well to verify the privacy-preserving ability of other GAN-based models.  

\textbf{PPGAN:} Similar to DPGAN, PPGAN also adds a well-designed noise to the gradients of the discriminator during the training process but uses an optimized noise addition algorithm to keep a higher utility. Unlike the original PPGAN code designed to generate synthetic images, we configured our framework to generate synthetic medical records. In our approach, we used a WGAN architecture as the skeleton of the PPGAN model and built the sequential functions for the discriminator, generator, and the GAN engine itself. We then added the noise function to the discriminator and trained the model, generated the synthetic data, and post-processed the dataset as suggested by \cite{paper2}

\subsection{QoG \& Privacy Evaluation} 
The synthetic datasets were generated by the five different GAN-based models used for comparison: (1) simpleGAN, (2) medGAN, (3) DG, (4) DG trained with  DPGAN, and (5) PPGAN. To assess our QoG, Long Short-term Memory (LSTM) has been implemented.
We evaluated the model performance in terms of the \textit{F1 Score}, the harmonic mean of the model's precision and recall, and root-mean-square error (RMSE). The LSTM was trained for the original data (denoted as \textit{real} data), the combination of real and synthetic data, and the synthetically generated datasets. 

For our predictive modeling, we have prepared our train and test sub-datasets for four different scenarios: (1) Only the real data is used for training and testing; (2) Combining the real dataset with each of the generated datasets and splitting the data into test and training datasets (\textit{real+simpleGAN, real+medGAN, real+DG, real+\textit{DPGAN (DG)}, and real+PPGAN}); (3) Each generated dataset is used for training and the real dataset for testing; (4) Each generated dataset is used for training and testing.

For the evaluation of the quality of the synthetic label values of \textit{user\_mmst} in the generated datasets, the autocorrelation coefficient was used to analyze and assess the similarities between the synthetically generated \textit{user\_mmst} and real \textit{user\_mmst} values for each GAN-based model. Also, the generated \textit{user\_mmst} sequence lengths with the real \textit{user\_mmst} sequence lengths as suggested by \cite{DoppleGANger} are compared. The privacy evaluation for the generated datasets and membership inference attacks against all of the datasets are performed using the \textit{Tensor Flow Privacy} framework. The model takes six parameters as input to initialize the attack, namely \textit{train \& test} labels, \textit{train \& test} losses, and, \textit{train \& test} logits (predictions). It also offers a data-slicing mechanism where separate attacks are executed on various chunks of the dataset. This can be useful to determine the robustness of specific sequences of the dataset. Also among various types of offered attacks, we empirically found that \textit{Logistic Regression} and the \textit{Threshold Attack} achieved the best performance on our data.

\section{Results} \label{results}

\begin{figure*}[ht!]
\begin{multicols}{2}
    \includegraphics[width=9cm, height=9cm]{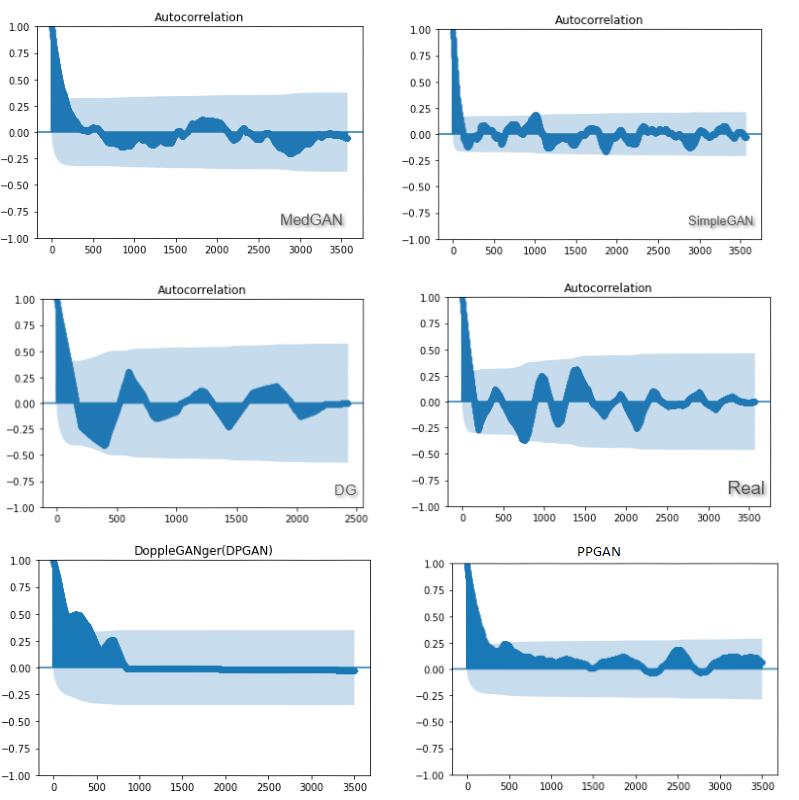}\par
    \caption{Autocorrelation of \textit{user\_mmst} values of the generated datasets compared to the real dataset.}
    \label{scf}
    \includegraphics[width=9cm, height=9cm]{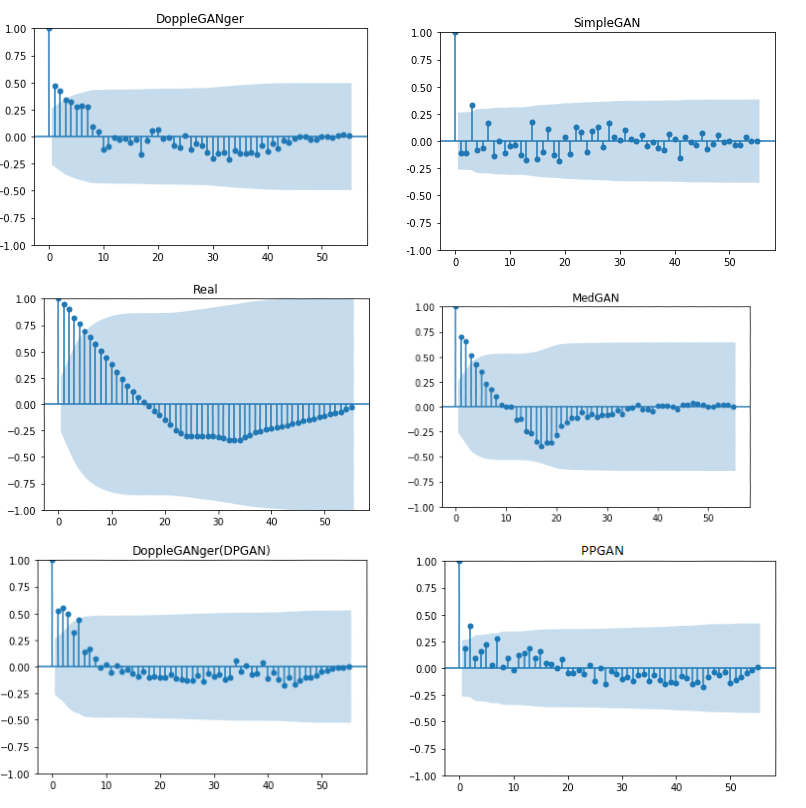}\par
    \caption{Autocorrelation of \textit{user\_mmst} values for the averaged generated datasets compared to the averaged real dataset.}
    \label{scf1}
    \end{multicols}
\label{fig}
\end{figure*}

\textbf{Predictive modeling:} \label{pm} Once our models have reached an optimal prediction accuracy, the models were compared concerning the \textit{F1 score} and \textit{RMSE} to determine the datasets which yield the best model performances for LSTM. In Table \ref{predtable} it is visible that the LSTM model performs best on the real dataset. However, the dataset generated by PPGAN combined with the real dataset yields an acceptable performance with a relatively low \textit{RMSE} value, followed by DG and PPGAN. From these findings, we can infer that the generated data by DG and PPGAN have less destructive effects when combined with the real data.

\begin{table}[htbp]
    \caption{LSTM model performance including RMSE and F1 scores of different datasets and the dataset combinations. }
    \begin{center}
    \begin{tabular}{lrr}
        \hline
        LSTM& RMSE& F1 Score  \\
        \hline
        Real & \textbf{0.01} & \textbf{0.81}\\
        Real+DG  & 0.28  & 0.36 \\ 
        Real+MedGAN & 1.24  & 0.16\\
        Real+SimpleGAN & 0.23  & 0.34\\
        Real+DPGAN (DG) &  \textbf{0.01} & \textbf{0.54}\\
        Real+PPGAN &  \textbf{0.25} &  \textbf{0.69}\\
        Train: DG - Test: Real & 0.61  & 0.32\\
        Train: MedGAN - Test: Real & 2.78 & 0.11\\
        Train: SimpleGAN - Test: Real & 0.12 & 0.01\\
        MedGAN & 5.90 &  0.19\\
        DG & \textbf{0.29}  &\textbf{0.57}\\
        SimpleGAN & 0.83 &  0.32 \\
        DPGAN (DG) & 3.31 &  0.10\\
        PPGAN &  \textbf{0.29} &   \textbf{0.54}\\
        \hline
    \end{tabular}
    \label{predtable}
    \end{center}
\end{table}

\textbf{Autocorrelation:} In Figure \ref{scf} the autocorrelation of the labeled data (\textit{user\_mmst}) in the real dataset and all the generated datasets are visualized to compare the underlying distribution of the real labels versus the generated labels. We can observe an almost sharp interval for the real and DG datasets at the beginning of the graph with convergence towards the higher lags. Whereas the graph in the case of medGAN and simpleGAN starts with an early convergence after a drastic drop at the beginning. PPGAN's graph, on the other hand, has a gradual convergence but it does not have minus values at the beginning of the graph unlike DG and the real datasets. Additionally, DPGAN's graph does not resemble the fluctuating pattern of the real data. Figure \ref{scf1} shows the same autocorrelation graphs but on the averaged dataset where all the columns per patient is averaged to have a single row representing each patient's data. In this case, more resemblance between the real and medGAN data is visible for the averaged datasets. \textit{DG's} pattern follows a similar pattern by gradually dropping from positive to negative and then raising back towards 0, except for a few cases around index 20 on the \textit{x} axis. The autocorrelation of both DPGAN and PPGAN for the averaged datasets looks relatively similar where none of the models exactly mimic the real dataset's pattern. Overall, medGAN, DG, and PPGAN show more similar autocorrelations to the real \textit{user\_mmst} values than the other models which is an indication of a higher QoG of these datasets. 

\textbf{Sequence length:} To further analyze the precision of the \textit{user\_mmst} values in the generated datasets, the length of the \textit{user\_mmst} columns in each sequence of the datasets were compared to the real dataset. We observed that the \textit{user\_mmst} values generated by medGAN and simpleGAN, follow a random Gaussian distribution whereas DG and PPGAN have a scattered distribution with more resemblance to the real dataset. Furthermore, DPGAN has faced a mode collapse situation with more than 90\% of its labeling values having a similar value. This might also suggest that the LSTM model has treated the values generated by the DPGAN(DG) as outliers (recall \ref{pm}. In summary, PPGAN has not fully captured the underlying distribution of the labels in the real data, but it had a better performance than the rest of the datasets (Figures were omitted due to brevity).

\textbf{Dataset Distribution Analysis:} Finally, the mean and standard deviation (STD) values in each dataset were generated for all columns and compared to the real dataset to assess the quality of the generated features. Hereby, We used heatmaps to visually compare the distribution of the features in each dataset and we found that PPGAN, DG, and medGAN had very similar feature distribution to the real dataset whereas DPGAN had the least similarity to the real dataset. These findings support the previously presented results (Figures were omitted due to brevity).

\textbf{Privacy evaluation}
The performance results of the two different attack types from the \textit{TensorFlow Privacy} framework are represented in Table \ref{attacktable}. The performance is denoted in terms of the \textit{Area Under the Curve} (AUC) value and an \textit{Attacker Advantage} (AA) value. The AUC values range from 0.5 to 1 with 0.5 indicating a random guess of whether the membership of a data sample can be inferred, and 1 meaning guaranteed membership inference. The AA denotes an attacker success probability ranging from 0 (0\%) to 1 (100\%).

\begin{table}[htbp]
\caption{Membership inference attack: logistic regression (LR), threshold attack (TA).}
\begin{center}
\begin{tabular}{ccccc}
\hline
Dataset & Percentile & Attack  & AUC & Attacker  \\
 & & Type & & Advantage \\
\hline
\multicolumn{1}{l}{Real}  &20-30  & LR  & 0.73 & 0.63 \\ 
\multicolumn{1}{l}{MedGAN} & 90-100 & LR & 0.65 & 0.30\\
\multicolumn{1}{l}{SimpleGAN} & 20-30 & LR & 0.67 & 0.41\\
\multicolumn{1}{l}{DG} & 60-70 & LR & 0.61 & 0.33\\
\multicolumn{1}{l}{DPGAN (DG)} & \textbf{20-30}  & \textbf{TA} &\textbf{0.73} & \textbf{0.55}\\
\multicolumn{1}{l}{PPGAN} & \textbf{80-90}  & \textbf{TA} & \textbf{0.55} &\textbf{0.09}\\
\hline
\end{tabular}
\label{attacktable}
\end{center}
\end{table}

Table \ref{attacktable} shows the attack results where PPGAN has yielded a 0.55  AUC score and only 0.09 AA which outperforms the rest of the dataset. whereas the \textit{DP\_DG} dataset has the same AUC value as the real data and a lower attacker advantage of 0.55 than the real data but still, has a relatively weaker performance compared to PPGAN.

\section{Discussion} \label{discussion}
%Limited access to medical records in research has led to the advancement of privacy-preserving data generation technologies \cite{medGAN}. The trade-off between QoG and \textit{differential privacy} is challenging to maintain as the more privacy-preserving a framework is, the more QoG has to be compromised.  
In this work, various analyses have been implemented to determine synthetic medical data generation's QoG and privacy aspects. The analysis of the QoG has been divided into three parts: (1) the predictive modeling analysis revealed a good performance of PPGAN combined with the real dataset achieving an F1 score of $0.69$ and also a low RMSE value of $0.25$. The second-best performance belonged to DG achieving an F1 score of $0.57$ and a very low RMSE value of $0.01$. (2) however, for the autocorrelation analysis medGAN showed the highest similarity with the original dataset, followed by DG and PPGAN. (3) When comparing the different models according to the sequence length the distribution of simpleGAN, medGAN, and PPGAN showed the highest similarity with the original distribution of the \textit{user\_mmst} values Although PPGAN appeared to be the most convenient model among all the QoG categories, a particular model could not be determined as significantly outperforming other models in all three aspects of determining the QoG.
When comparing the different GAN-based approaches in terms of privacy, PPGAN clearly outperforms the other approaches when applying membership inference attacks. Hereby, PPGAN achieved a very low AUC value of $0.55$ indicating a random inference attack. Also, the AA value was very low ($0.09$). Thus, PPGAN has shown a lower risk of data leakage while at the same time keeping an acceptable QoG
%\footnote{The source code for PPGAN and the adjustments for this use case will be fully made available after the blind review}.

Achieving high predictive model accuracy - despite having paramount importance in the case of medical data - is still an open challenge given an insufficient amount of data. additionally, there are no existing standard thresholds when accuracy is ultimately judged as sufficient. Hence, achieving higher model accuracy needs to be addressed in future work regarding the generation of synthetic medical data.

\section{Conclusion} \label{conclusion}
%The demand for high-quality, privacy-preserving datasets in various research and industrial domains is rising. The main challenge concerning synthetic data generation is to produce synthetic datasets with the least data leakage possibility while maintaining a high QoG in order to improve QoE for various applications. 

In this work, we have used state-of-the-art, GAN-based, synthetic data generation frameworks to generate synthetic medical records from a dataset containing time series medical records about dementia patients. The GAN-based models, namely simpleGAN, medGAN, DG, DPGAN, and PPGAN have been compared based on three aspects of predictive modeling, autocorrelation, and sequence length wherein PPGAN has appeared as the dominant model overall categories. In terms of privacy aspects, PPGAN outperformed other models when subjected to membership inference attacks and showed a low risk of re-identification of patients. Thus, PPGAN is recommended for generating synthetic medical records, keeping an acceptable balance between privacy preservation and QoG. Since this analysis has been conducted on a single dataset, our claim only applies to scenarios in which similar datasets are used, plus an optimistic anticipation of gaining comparable results on different datasets.
Overall, the analysis shows that there is room for future research on improving the QoG and privacy aspects of the state-of-the-art GAN-based models for synthetic data generation. Furthermore, by using suggested models, it is possible to develop services and platforms that provide a high QoE while guaranteeing privacy preservation.

\bibliographystyle{IEEEtran}
\bibliography{conference_101719.bib}

\end{document}